\documentclass[]{spie}  

\usepackage{amsmath,amsfonts,amssymb}
\usepackage{subcaption}
\usepackage{graphicx}
\usepackage{cite} 
\usepackage{epsfig}
\usepackage{amsmath}
\usepackage{nccmath}
\usepackage{amssymb}
\usepackage{mwe}
\usepackage{acro}
\usepackage{amssymb}
\usepackage{xcolor,colortbl}
\usepackage{tabularx}
\usepackage{relsize}
\usepackage{pifont}
\usepackage{booktabs} 
\usepackage{multirow}
\usepackage{multicol}
\usepackage{adjustbox}
\usepackage{float}
\usepackage{graphicx}
\usepackage{makecell}
\usepackage{tabu}
\usepackage[colorlinks=true, allcolors=blue]{hyperref}
\usepackage[capitalize]{cleveref}
\usepackage{tabularray}

\usepackage{booktabs}

\title{MagNet: Multi-Level Attention Graph Network for Predicting High-Resolution Spatial Transcriptomics}

\author[a]{Junchao Zhu}
\author[b]{Ruining Deng}
\author[a]{Tianyuan Yao}
\author[a]{Juming Xiong}
\author[a]{Chongyu Qu}
\author[a]{Junlin Guo}
\author[a]{Siqi Lu}
\author[c]{Yucheng Tang}
\author[c]{Daguang Xu}
\author[d]{Mengmeng Yin}
\author[d]{Yu Wang}
\author[d]{Shilin Zhao}
\author[e]{Yaohong Wang}
\author[d]{Haichun Yang}
\author[a,d]{Yuankai Huo*}
\affil[a]{Vanderbilt University, Nashville, TN, USA}
\affil[b]{Weill Cornell Medicine, NY, USA}
\affil[c]{NVIDIA, WA, USA}
\affil[d]{Vanderbilt University Medical Center, Nashville, TN, USA}
\affil[e]{UT MD Anderson Cancer Center, TX, USA}

\authorinfo{Corresponding author: Yuankai Huo: E-mail: yuankai.huo@vanderbilt.edu}

\begin{document} 
\maketitle

\begin{abstract}
The rapid development of spatial transcriptomics (ST) offers new opportunities to explore the gene expression patterns within the spatial microenvironment. Current research integrates pathological images to infer gene expression, addressing the high costs and time-consuming processes to generate spatial transcriptomics data. However, as spatial transcriptomics resolution continues to improve, existing methods remain primarily focused on gene expression prediction at low-resolution (55~$\mu$m) spot levels. These methods face significant challenges, especially the information bottleneck, when they are applied to high-resolution (8~$\mu$m) HD data. To bridge this gap, this paper introduces MagNet, a multi-level attention graph network designed for accurate prediction of high-resolution HD data. MagNet employs cross-attention layers to integrate features from multi-resolution image patches hierarchically and utilizes a GAT-Transformer module to aggregate neighborhood information. By integrating multilevel features, MagNet overcomes the limitations posed by low-resolution inputs in predicting high-resolution gene expression. We systematically evaluated MagNet and existing ST prediction models on both a private spatial transcriptomics dataset and a public dataset at three different resolution levels. The results demonstrate that MagNet achieves state-of-the-art performance at both spot level and high-resolution bin levels, providing a novel methodology and benchmark for future research and applications in high-resolution HD-level spatial transcriptomics. Code is available at \href{https://github.com/Junchao-Zhu/MagNet}{https://github.com/Junchao-Zhu/MagNet}.

\end{abstract}


\section{Introduction}

Spatial transcriptomics (ST) provides a novel view for correlating pathological tissue structures with their spatial gene expression patterns~\cite{burgess2019spatial, asp2019spatiotemporal,he2020integrating}. This approach advances the development of effective treatment strategies~\cite{asp2020spatially}. Studies have demonstrated a strong correlation between features of pathological images and their gene expression patterns~\cite{badea2020identifying}. In recent years, the widespread application of deep learning methods in medical image analysis~\cite{he2020integrating, zhou2018unet++} has made it possible to predict gene expression from broadly accessible and affordable whole-slide images (WSIs).

Currently, several studies have employed methods such as convolutional neural networks (CNNs)~\cite{he2020integrating,xie2024spatially,yang2023exemplar} and graph neural networks (GNNs)~\cite{pang2021leveraging, zeng2022spatial, jia2024thitogene} to predict spatial transcriptomic expression at the spot level with low resolution. These approaches exploit spatial dependencies~\cite{zeng2022spatial, pang2021leveraging} and image similarities~\cite{xie2024spatially,yang2023exemplar} inherent in pathological images, thus integrating information to optimize the fusion of image features. Such advances address the challenges of scarce high-quality spatial transcriptomic data and the high cost of acquisition.

\begin{figure*}
    \centering
    \includegraphics[width=0.85\linewidth]{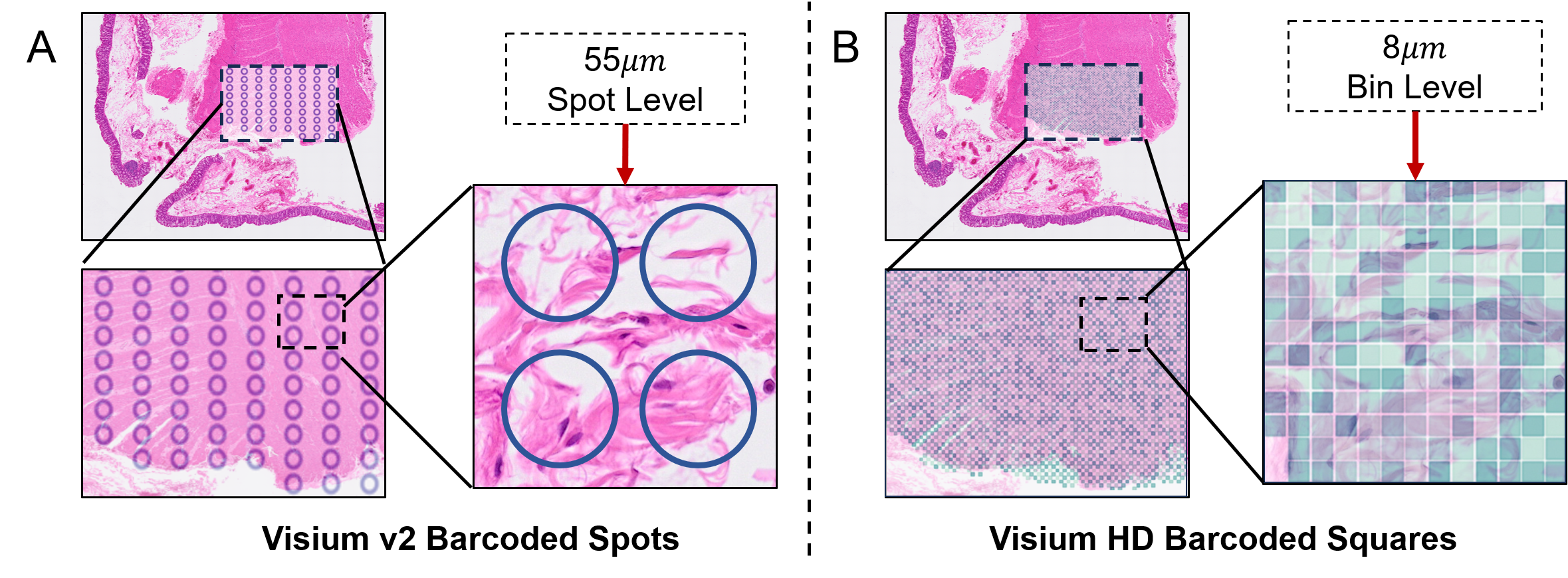}
    \caption{\textbf{Spatial transcriptomics data at different resolutions.} (A) Traditional low-resolution 10X Visium v2 barcoded spots, where spots are discretely distributed with a diameter of 55~$\mu$m. (B) Current high-resolution 10X Visium HD barcoded squares, where bins are densely distributed with a diameter of 8~$\mu$m.}
    \label{fig:Demo}
\end{figure*}

Continuous advancements in ST sequencing technology~\cite{staahl2016visualization, wang2018three, eng2019transcriptome} have significantly improved the resolution of existing ST data, as is shown in Figure~\ref{fig:Demo}, which has progressed from the initial 55~$\mu$m spots to higher resolutions, such as HD data with bin diameters of 8~$\mu$m or even 2~$\mu$m. Such advancement enables a more comprehensive analysis of the relationship between pathological tissues and gene expression at the single-cell level~\cite{zhu2024stie,benjamin2024multiscale,oliveira2024characterization,janesick2023high}. However, current deep-learning methods face an information bottleneck when dealing with high-resolution HD data~\cite{tishby2015deep}. Specifically, the limited information from low-resolution input images is insufficient to effectively support the prediction of high-dimensional gene expression. The features extracted by these models may lack the complexity required to represent the intricate details of high-resolution, high-dimensional gene expression data.

To address this issue, this paper proposes MagNet, a Multi-Level Attention Graph Network designed for accurate prediction of high-resolution HD data. MagNet integrates information across multiple resolutions, including the bin, spot, and region levels, through cross-attention layers. MagNet also extracts and combines features from neighboring regions with Graph Attention Network (GAT) and Transformer layers. Thus, our proposed framework overcomes the information bottleneck posed by low-resolution inputs when predicting high-resolution, high-dimensional gene expression by efficient extraction and integration of multisource and multilevel features. Furthermore, the model incorporates cross-resolution constraints on gene expression within the same region, further enhancing its performance in HD gene expression prediction.
Our contributions can be summarized in three aspects:

\textbullet\ We present MagNet, a Multi-Level Attention Graph Network designed for accurate prediction of high-resolution HD data. To our knowledge, it is the first model dedicated to HD-level gene expression prediction.

\textbullet\ Our proposed framework leverages cross-attention layers and GAT-Transformer blocks to effectively extract and integrate multi-source and multi-level features, tackling the information bottleneck of low-resolution inputs in predicting high-resolution ST expression.

\textbullet\ We provide our model as an open-source tool, benchmarking and providing a systematic evaluation on a privately-collected kidney HD ST dataset and a public colorectal cancer HD ST dataset. 
\section{Method}

\begin{figure*}[htb]
    \centering
    \includegraphics[width=\linewidth]{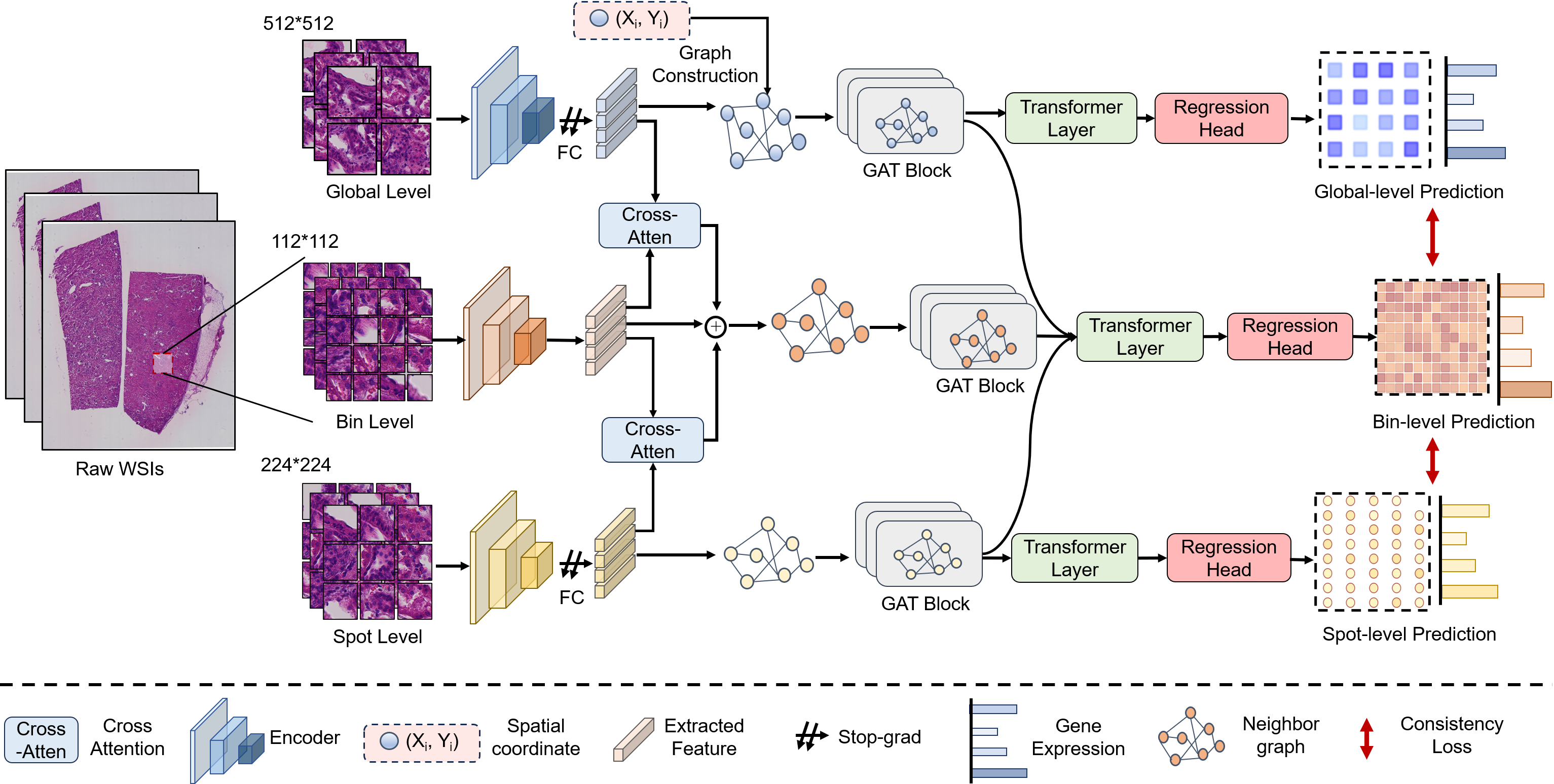}
    \caption{\textbf{The network structure of the proposed MagNet.} MagNet utilizes cross-attention layers to integrate features extracted from multi-resolution patches. Additionally, it incorporates a GAT-Transformer block to aggregate neighborhood information while leveraging spatial relationships. The predictions for each resolution level are then independently generated by a regression head.}
    \label{fig:framework}
\end{figure*}

\subsection{Unified Cross-Resolution Feature Aggregation}
We cropped patches at the bin, spot, and region levels for each bin $i$, denoted as $i_b$, $i_s$ and $i_r$. Features of these patches, represented as $f_b$, $f_s$ and $f_r$, are extracted by a pre-trained ResNet50~\cite{he2016deep}. We adopt the strategy proposed by TRIPLEX~\cite{chung2024accurate} that freezes the encoder parameters for the spot and region levels while updating only the bin-level encoder to minimize computational overhead.

To refine the representation of $f_b$, the features of other resolutions are treated as the key matrix (K) and the value matrix (V), with $f_b$ acting as the query matrix (Q). A cross-attention layer is used to effectively merge the features of $f_s$ and $f_r$ into $f_b$. Thus, the fused feature $f'_b$ is formulated as:
\begin{equation}
f'_b = \text{softmax} \left( \frac{f_b f_i^T}{\sqrt{d}} \right) f_i, \quad i=s, r
\end{equation}
where $\sqrt{d}$ is a scaling factor. Finally, by concatenating the features from all three levels, the fused multi-level feature $F$ is obtained for use in subsequent processes.

\subsection{Spatial-Guided Graph Integration Block}
To exploit the spatial relationship of pathological images, we propose a spatially-guided graph integration block that integrates GAT and transformer layers. The connections between bins are first established by calculating the weight $e_{ij}$ between any two nodes $i$ and $j$ using the Euclidean distance. The top-$k$ lowest $e_{ij}$ values are selected to establish connections within the whole-slide image. The constructed graph is then fed into the spatial-guided graph integration block for further processing.

Subsequently, after rounds of graph attention convolution, the processed feature $F_m^{i}$ for each $i_b$, $i_s$ and $i_r$ is formulated as follows:

\begin{equation}
\mathbf{F}_m^{i} = \bigg\Vert_{k=1}^K \sigma \left( \sum_{j \in \mathcal{N}(i)} \alpha_{ij}^k \mathbf{W}^k \mathbf{f}_m ^{j}\right), \{m|b,s,r\}
\end{equation}
where $\mathcal{N}(i)$ denotes the set of adjacent nodes, $\bigg\Vert$ represents concatenation operation, $\sigma$ is the activation function, $\alpha_{ij}^k$ is the weight of the $k$-th attention head, and $\mathbf{W}^k$ is a linear transformation matrix determined by the connections between nodes. 

A Transformer layer is used for adaptive aggregation of neighborhood information from each round, thus enhancing the representation of features. Finally, the regression head generates gene expression predictions for each level separately, denoted as $p_b$, $p_s$, and $p_r$.

\subsection{Loss Function}
To exploit the mutual consistency among multilevel information, we designed a hybrid loss function comprising prediction loss $L_p$ and consistency loss $L_c$ to optimize the model learning process. The prediction loss primarily focuses on minimizing the discrepancies between the model's predictions and the ground truth at each resolution level. For the prediction task at bin level, we employ Mean Squared Error (MSE) and Pearson Correlation Coefficient loss (P) to evaluate the model's performance. To avoid introducing additional noise, only PCC loss is utilized to assess the model's performance at the spot and region levels. Hence, the prediction loss is formulated as:
\begin{equation}
 L_p = MSE(p_b, y_b) + \sum_{i=b,s,r} \lambda_i \cdot P(p_i, y_i) 
\end{equation} %
Here, $b$, $s$, and $r$ represent the bin, spot, and region levels, respectively. $p_i$ and $y_i$ denote the prediction of the model and its corresponding ground truth, while $\lambda_i$ is a hyperparameter used to balance the PCC loss at different resolution levels.

Since patches at different resolutions within the same region exhibit similar trends in gene expression, we employ PCC loss to constrain the differences between bin-level predictions and those at other levels. The consistency loss $L_c$ is defined as:
\begin{equation}
 L_c = \lambda_1 \cdot P(p_b, p_s) + \lambda_2 \cdot P(p_b, p_r) 
\end{equation} 
Thus, the overall loss of the model $L$ is defined as:
\begin{equation} 
L = \gamma_1 \cdot L_p + \gamma_2 \cdot L_c 
\end{equation}
Here, $\gamma_1$ and $\gamma_2$ are hyperparameters used to balance the two types of losses, and they are set to 1 and 0.25 in the subsequent experiments.

\section{Data and Experiment}
\textbf{Dataset.} We benchmarked our MagNet and other baseline models on a privately collected kidney pathology dataset (VUMC) and a publicly available colorectal cancer (CRC) dataset~\cite{oliveira2024characterization}. We conducted four-fold cross-validation at the WSI level. Our in-house dataset contains 12 HD ST samples with three resolutions: 2~$\mu$m, 8~$\mu$m, and 16~$\mu$m, where 1px in the WSI corresponds to 0.25 $\mu$m of real tissue. The CRC dataset consists of four samples with a single-layer section, including two CRC tissues and two adjacent normal tissues. The process has been approved by Institutional Review Board (IRB).

\noindent\textbf{Data Preprocessing.}
6,000 bins were randomly selected for each WSI, and 112$\times$112 pixel patches centered at 8~$\mu$m and 16~$\mu$m bins were cropped. At the spot and region levels, patches with diameters of 224 and 512 pixels were extracted across the WSI, with their gene expressions aggregated from bin-level data. 2,500 spot-level patches per WSI were selected for training and testing. Patch pairing across levels was based on the distance between the coordinates in different resolutions. We follow the method proposed in ST-Net~\cite{he2016deep} and select the top 250 genes with the highest average expression levels of more than 20,000 original genes for prediction. Gene expression values were normalized using the approach introduced in TRIPLEX~\cite{chung2024accurate}, which involves proportional normalization followed by a log transformation.

\begin{table*}[t]
\centering
\caption{\textbf{Quantitative comparisons across different datasets.} 
The best performance is highlighted in \textbf{bold}, where we can observe that  \texttt{MagNet} outperforms the state-of-the-art in multiple resolutions.}
\resizebox{\textwidth}{!}{
\begin{tblr}{
  cells = {c},
  cell{1}{1} = {r=2}{},
  cell{1}{2} = {r=2}{},
  cell{1}{3} = {c=3}{},
  cell{1}{6} = {c=3}{},
  cell{3}{1} = {r=6}{},
  cell{9}{1} = {r=6}{},
  cell{15}{1} = {r=6}{},
  vline{2-3,6} = {3-8,9-14,15-20}{},
  vline{3,6} = {4-8,10-14,16-20}{},
  hline{1,3,9,15,21} = {-}{},
  hline{2} = {3-8}{},
}
Resolution & Model     & VUMC (in-house dataset) &                      &                      & CRC~\cite{oliveira2024characterization}   &                      &                      \\
           &           & MSE                     & MAE                  & PCC                  & MSE                  & MAE                  & PCC                  \\
8$\mu$m /112px    & ST-Net    & 0.193±0.004             & 0.388±0.009          & 0.226±0.040          & 0.292±0.076          & 0.402±0.084          & 0.527±0.155          \\
           & EGN       & 0.048±0.011             & 0.134±0.020          & 0.157±0.024          & 0.409±0.164          & 0.508±0.139          & 0.511±0.152          \\
           & HisToGene & 0.105±0.007             & 0.241±0.006          & 0.109±0.018          & 0.311±0.088          & 0.419±0.075          & 0.451±0.128          \\
           & BLEEP     & 0.063±0.006             & 0.163±0.009          & 0.199±0.052          & 0.348±0.041          & 0.440±0.0361         & 0.475±0.1379         \\
           & His2ST    & 0.140±0.019             & 0.358±0.026          & 0.175±0.033          & 0.287±0.113          & 0.4041±0.109         & 0.537±0.165          \\
           & \textbf{MagNet(Ours)}   & \textbf{0.048±0.008}    & \textbf{0.109±0.008} & \textbf{0.278±0.042} & \textbf{0.271±0.054} & \textbf{0.375±0.053} & \textbf{0.541±0.167} \\
16$\mu$m /112px   & ST-Net    & 0.288±0.007             & 0.420±0.027          & 0.364±0.0539         & 0.661±0.239          & 0.632±0.146          & 0.560±0.151          \\
           & EGN       & 0.149±0.037             & 0.302±0.06           & 0.308±0.037          & 0.740±0.0241         & 0.677±0.013          & 0.552±0.014          \\
           & HisToGene & 0.204±0.045             & 0.380±0.052          & 0.243±0.035          & 0.660±0.176          & 0.6368±0.099         & 0.522±0.136          \\
           & BLEEP     & 0.174±0.029             & 0.290±0.031          & 0.317±0.058          & 0.673±0.161          & 0.625±0.088          & 0.504±0.123          \\
           & His2ST    & 0.224±0.044             & 0.427±0.049          & 0.330±0.046          & 0.610±0.168          & 0.611±0.103          & 0.562±0.152          \\
           & \textbf{MagNet(Ours)}  & \textbf{0.127±0.024}    & \textbf{0.228±0.034} & \textbf{0.378±0.057} & \textbf{0.564±0.184} & \textbf{0.581±0.114} & \textbf{0.574±0.154} \\
55$\mu$m /224px   & ST-Net    & 0.442±0.036             & 0.549±0.019          & 0.609±0.059          & 0.767±0.203          & 0.652±0.086          & 0.649±0.080          \\
           & EGN       & 0.355±0.030             & 0.471±0.010          & 0.601±0.0561         & 0.778±0.229          & 0.651±0.105          & \textbf{0.674±0.071} \\
           & HisToGene & 0.403±0.028             & 0.517±0.017          & 0.596±0.058          & 0.702±0.173          & 0.622±0.074          & 0.663±0.067          \\
           & BLEEP     & 0.339±0.026             & 0.467±0.017          & 0.576±0.049          & 0.717±0.112          & 0.623±0.044          & 0.667±0.043          \\
           & His2ST    & 0.327±0.021             & 0.459±0.013          & 0.601±0.058          & 0.813±0.199          & 0.673±0.089          & 0.673±0.065          \\
           & \textbf{MagNet(Ours)}  & \textbf{0.324±0.044}    & \textbf{0.458±0.030} & \textbf{0.611±0.082} & \textbf{0.688±0.149} & \textbf{0.612±0.069} & 0.670±0.059          
\end{tblr}
    }
        \label{table:comparison}
\end{table*}

\noindent\textbf{Compared Methods and Evaluation Metrics.} MagNet was benchmarked against current ST counterparts, including spatial-aware methods HisToGene~\cite{pang2021leveraging} and His2ST~\cite{zeng2022spatial}, similarity-based strategy BLEEP~\cite{xie2024spatially} and EGN~\cite{yang2023exemplar}, and the classic approach ST-Net~\cite{he2020integrating}. We used the officially released code published along with the papers for all of the methods. The Pearson correlation coefficient (PCC), mean squared error (MSE), and mean absolute error (MAE) are used to evaluate the performance of the models comprehensively.

\noindent\textbf{Experiment Setting and Implementation.} 
Experiments were conducted on NVIDIA RTX A6000 GPU cards. The SGD optimizer was utilized, with momentum set to 0.9 and a weight decay of $10^{-4}$. An initial learning rate of $10^{-4}$ was applied, which followed a cosine decay schedule, reducing it to 0.01 of its starting value throughout training. All models are trained to converge. We employed a batch size of 256 for training and fine-tuned the hyperparameters $\lambda_1$, $\lambda_2$, $\lambda_b$, $\lambda_s$, and $\lambda_r$ in our hybrid loss function to values of 0.1, 0.1, 0.8, 0.25, and 0.25, respectively. For graph construction, the top-$k$ value was fixed at 8. We select 8~$\mu$m and 16~$\mu$m bins as the target HD resolution to predict, due to the extremely low gene expression amount in 2~$\mu$m bins. During spot-level experiments, we freeze the encoder parameters of the bin and region levels and update the spot level instead. 

\section{Results}
\subsection{Cross-Validation Evaluation}
We conducted four-fold validation on the WSI level to validate and benchmark MagNet and SOTAs on the two HD datasets. Table~\ref{table:comparison} summarizes quantitative comparisons of various baselines across different datasets and resolutions. Our proposed MagNet consistently outperforms existing methods in almost all metrics, with its superiority particularly evident at HD high-resolution levels. Taking the 8~$\mu$m prediction task in our VUMC dataset as an example, MagNet achieved MSE, MAE, and PCC values of 0.048±0.008, 0.109±0.008, and 0.278±0.042, respectively, significantly surpassing the results of other methods, such as BLEEP, which reported values of 0.063±0.006, 0.163±0.009, and 0.199±0.052.

These findings demonstrate the capability of MagNet to effectively address the information bottleneck inherent in high-resolution gene prediction tasks. By efficiently integrating and leveraging multi-source and multi-level information, MagNet overcomes the performance limitations caused by constrained data and substantially enhances prediction accuracy for high-resolution HD data. Furthermore, the relatively low standard deviation observed among all metrics during cross-validation highlights the method's robustness and stability, underscoring its reliability for practical clinical applications.

\subsection{Pivotal Gene Expression Prediction}
We evaluated the clinical applicability of various baselines by analyzing the predictive performance of the key biomarker SGPP1 in our kidney dataset. SGPP1 and its associated pathways play a critical role in kidney health and disease, with direct implications for conditions such as acute kidney injury and fibrotic kidney diseases~\cite{drexler2021sphingosine, keller2024factors, lovric2017mutations}.

\begin{figure*}[t]
    \centering
    \includegraphics[width=\linewidth]{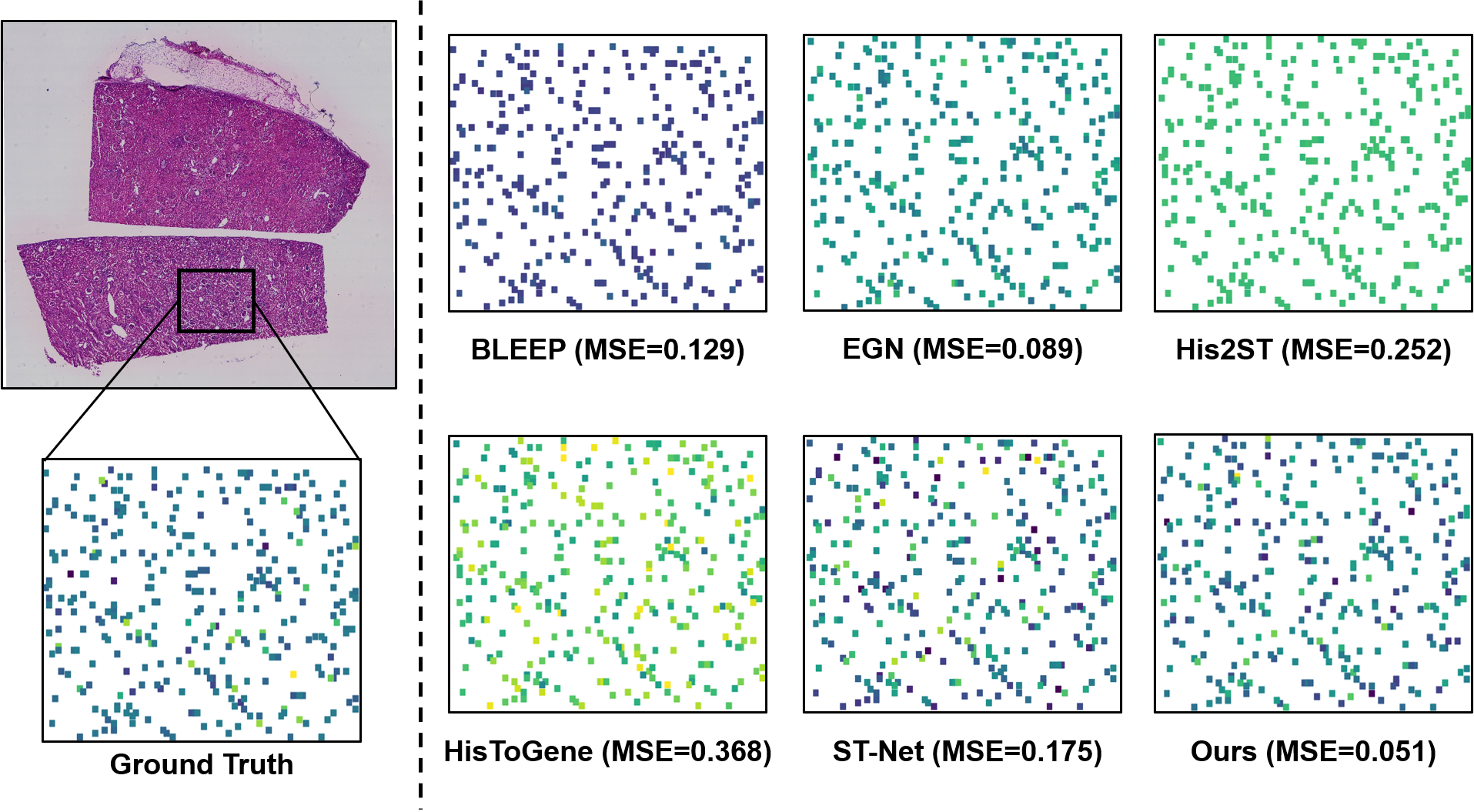}
    \caption{\textbf{Qualitative comparison for pivotal gene expression prediction.} }
    \label{fig:key_gene}
\end{figure*}

Figure~\ref{fig:key_gene} illustrates the predictive performance of different models for the SGPP1 gene. Compared with other baseline models, our proposed MagNet achieved the best MSE of 0.051. By deeply integrating and leveraging multi-level information, MagNet captures the spatial distribution of key gene expressions in pathological tissues with higher resolution, providing more detailed predictions for subsequent diagnoses and demonstrating strong potential for clinical applications.

\subsection{Ablation Study}
We conducted a detailed ablation study to evaluate the effectiveness of each functional block, as is summarized in Table~\ref {table:ablation}. The results demonstrate that incorporation of the GAT-Transformer block and multi-resolution information effectively compensates for the limited information in the original bin-level data, thus significantly increasing the PCC by 0.079 in our dataset and by 0.026 in the CRC dataset. 

Additionally, introducing a consistency loss enhances the synergy of multi-resolution information by leveraging the mutual constraints of gene expression across different resolutions within the same region, thereby facilitating more effective learning of high-resolution features and further improving the model's performance. To conclude, the benefits of each block are mutually exclusive and synergistic, allowing MagNet to achieve optimal results when integrating all modules.

\begin{table}[t]
\centering
\caption{\textbf{Ablation study for functional blocks in MagNet.} The benefits from each designed block are orthonormal, while MagNet achieves optimal results when integrating all modules.}
\resizebox{\textwidth}{!}{
\begin{tblr}{
  row{1} = {c},
  row{2} = {c},
  cell{1}{1} = {r=2}{},
  cell{1}{2} = {c=3}{},
  cell{1}{5} = {c=3}{},
  cell{3}{2} = {c},
  cell{3}{3} = {c},
  cell{3}{4} = {c},
  cell{3}{5} = {c},
  cell{3}{6} = {c},
  cell{3}{7} = {c},
  cell{4}{2} = {c},
  cell{4}{3} = {c},
  cell{4}{4} = {c},
  cell{4}{5} = {c},
  cell{4}{6} = {c},
  cell{4}{7} = {c},
  cell{5}{2} = {c},
  cell{5}{3} = {c},
  cell{5}{4} = {c},
  cell{5}{5} = {c},
  cell{5}{6} = {c},
  cell{5}{7} = {c},
  cell{6}{2} = {c},
  cell{6}{3} = {c},
  cell{6}{4} = {c},
  cell{6}{5} = {c},
  cell{6}{6} = {c},
  cell{6}{7} = {c},
  cell{7}{2} = {c},
  cell{7}{3} = {c},
  cell{7}{4} = {c},
  cell{7}{5} = {c},
  cell{7}{6} = {c},
  cell{7}{7} = {c},
  hline{1,3,8} = {-}{},
  hline{2} = {2-7}{},
}
Functional Blocks                 & VUMC (in-house dataset) /16$\mu$m    &             &             & CRC~\cite{oliveira2024characterization}/16 $\mu$m    &             &             \\
                                  & MSE         & MAE         & PCC         & MSE         & MAE         & PCC         \\
w.o. GAT \&  Multi-resolution & 0.148±0.042 & 0.281±0.069 & 0.299±0.028 & 0.799±0.259 & 0.709±0.146 & 0.548±0.146 \\
w.o. GAT block      & 0.135±0.030 & 0.266±0.048 & 0.306±0.043 & 0.632±0.170 & 0.624±0.096 & 0.550±0.147 \\
w.o. Multi-resolution             & 0.133±0.030 & 0.260±0.051 & 0.323±0.044 & 0.634±0.175 & 0.628±0.111 & 0.563±0.152 \\
w.o. Consistency Loss             & 0.130±0.023 & 0.235±0.040 & 0.369±0.054 & 0.624±0.187 & 0.619±0.117 & 0.559±0.146 \\
w. All blocks       &  \textbf{0.127±0.024} & \textbf{0.228±0.034} & \textbf{0.378±0.057} & \textbf{0.564±0.184} & \textbf{0.581±0.114} & \textbf{0.574±0.154}
\end{tblr}   }
        \label{table:ablation}
\end{table}
\section{Conclusion}
We introduce a novel framework specifically tailored for high-resolution gene expression tasks. Our MagNet model integrates multi-level information and leverages spatial relationships derived from pathological images, effectively overcoming the input-information bottleneck in HD gene expression prediction. Consequently, MagNet can accurately capture gene expression patterns at an 8~$\mu$m single-cell resolution. In addition, we present the first systematic and comprehensive evaluation of HD-level spatial transcriptomics datasets. We benchmarked MagNet against current state-of-the-art methods on two HD datasets under three different resolution settings. Experimental results demonstrate that MagNet consistently achieves top-tier predictive performance across multiple resolutions in both datasets. By extending gene prediction from the spot level to the cellular scale, MagNet establishes a new paradigm and benchmark for future research in spatial transcriptomics.

\acknowledgments 
This research was supported by NIH R01DK135597(Huo), DoD HT9425-23-1-0003(HCY), NIH NIDDK DK56942 (ABF). This work was also supported by Vanderbilt Seed Success Grant, Vanderbilt Discovery Grant, and VISE Seed Grant. This project was supported by The Leona M. and Harry B. Helmsley Charitable Trust grant G-1903-03793 and G-2103-05128. This research was also supported by NIH grants R01EB033385, R01DK132338, REB017230, R01MH125931, and NSF 2040462. We extend gratitude to NVIDIA for their support by means of the NVIDIA hardware grant.

\bibliography{report} 
\bibliographystyle{spiebib} 

\appendix

\section{Gene Selection and Estimation}
To estimate the gene expression at the spot level and the region level, we aggregated the value of gene expression of 16 $\mu$m bins within their respective spot and region areas. This process can be defined as:

\begin{equation} y_s = \sum_{i \in S} y_i, \quad y_r = \sum_{i \in R} y_i \end{equation}

\begin{figure*}[t]
    \centering
    \includegraphics[width=\linewidth]{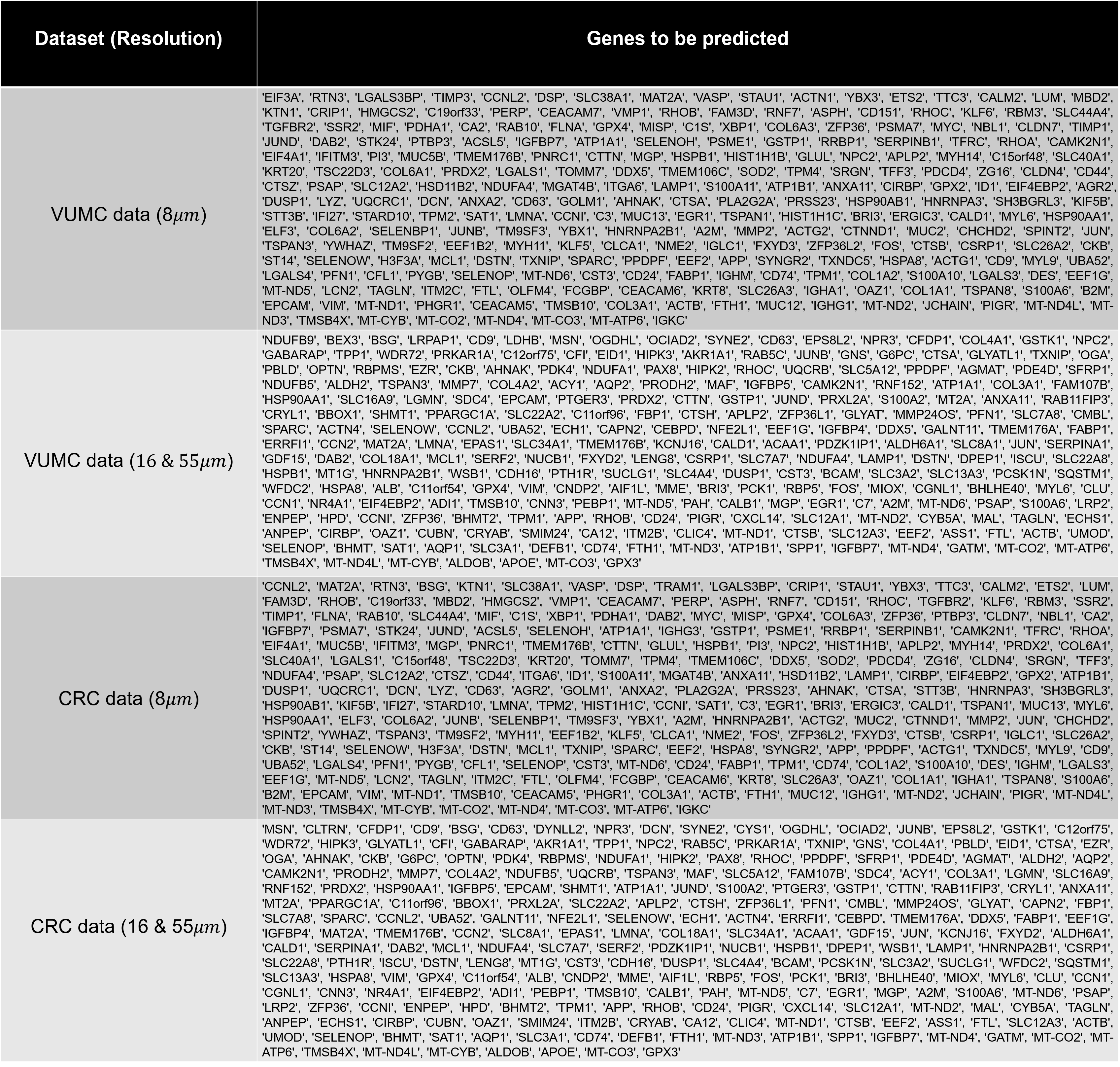}
    \caption{\textbf{Gene selection in each dataset and resolution.} }
    \label{fig:key_genes}
\end{figure*}

Here, $y_i$ denotes the gene expression value at the $i$-th bin, $S$ represents the set of bins within a specific spot, and $R$ denotes the set of bins within a certain area, thus ensuring the consistency of gene expression across multiple resolutions. The selected genes with the highest average expression for each dataset and resolution are presented in Figure.~\ref{fig:key_genes}

\end{document}